\documentclass[10pt]{cai}

\begin{document}
\def\conferenceyear{2025}
\volumeheader{38}{0}
\begin{center}

\title{
    Cardioformer: Advancing AI in ECG Analysis with Multi-Granularity Patching and ResNet
}
\maketitle

\thispagestyle{empty}

\begin{tabular}{cc}
Md Kamrujjaman Mobin\upstairs{1,*}, Md Saiful Islam\upstairs{1,2,*}, Sadik Al Barid\upstairs{1}, Md Masum\upstairs{1}
\\[0.25ex]
{\small \upstairs{1} Shahjalal University of Science and Technology, Sylhet, Bangladesh} \\
{\small \upstairs{2} Athabasca University, Canada} \\
\end{tabular}
  
\emails{
  \upstairs{*}kamrujjamanmobin123@gmail.com,
  \upstairs{*}sislam@athabascau.ca, nebir2002@gmail.com, masum-cse@sust.edu
}
\end{center}

\begin{abstract}

Electrocardiogram (ECG) classification is crucial for automated cardiac disease diagnosis, yet existing methods often struggle to capture local morphological details and long-range temporal dependencies simultaneously. To address these challenges, we propose Cardioformer, a novel multi-granularity hybrid model that integrates cross-channel patching, hierarchical residual learning, and a two-stage self-attention mechanism. Cardioformer first encodes multi-scale token embeddings to capture fine-grained local features and global contextual information and then selectively fuses these representations through intra- and inter-granularity self-attention. Extensive evaluations on three benchmark ECG datasets under subject-independent settings demonstrate that model consistently outperforms four state-of-the-art baselines. Our Cardioformer model achieves the AUROC of \textbf{96.34$\pm$\text{\tiny 0.11}}, \textbf{89.99$\pm$\text{\tiny 0.12}}, and \textbf{95.59$\pm$\text{\tiny 1.66}} in MIMIC-IV, PTB-XL and PTB dataset respectively outperforming PatchTST, Reformer, Transformer, and Medformer models. It also demonstrates strong cross-dataset generalization, achieving \textbf{49.18\%} AUROC on PTB and \textbf{68.41\%} on PTB-XL when trained on MIMIC-IV. These findings underscore the potential of Cardioformer to advance automated ECG analysis, paving the way for more accurate and robust cardiovascular disease diagnosis. We release the source code at \url{https://github.com/KMobin555/Cardioformer}.

\end{abstract}

\copyrightnotice

\vspace*{-0.2in}

\section{Introduction}

Cardiovascular diseases (CVDs) remain a leading global health challenge, responsible for approximately 17.9 million deaths annually \cite{who2025cardiovascular}. In the United States alone, CVDs claimed 941,652 lives in 2022 \cite{heart2025statistics}, and projections suggest that by 2050, the prevalence of these conditions could surge by up to 90\%, potentially resulting in 35.6 million deaths worldwide \cite{chong2024global}. These stark statistics underscore the imperative for early detection and accurate diagnosis in mitigating the impact of CVDs.

\noindent
\begin{minipage}{0.58\textwidth}
    ECG signals are vital for cardiac assessment but challenging to interpret due to the need to capture both local details and long-range dependencies. Traditional methods, including biomarker extraction, CNNs, and GNNs, achieve moderate success but struggle with the complex temporal dynamics of multi-channel ECG data. Recent advances in Transformer models have revolutionized time series analysis with self-attention \cite{vaswani2017attention}. While Autoformer \cite{wu2021autoformer} and Informer \cite{zhou2021informer} focus on single-channel timestamps, PatchTST \cite{nie2023timeseries}, Reformer \cite{kitaev2019reformer}, 
\end{minipage}
\hfill
\begin{minipage}{0.4\textwidth}
    \centering
    \includegraphics[width=\linewidth]{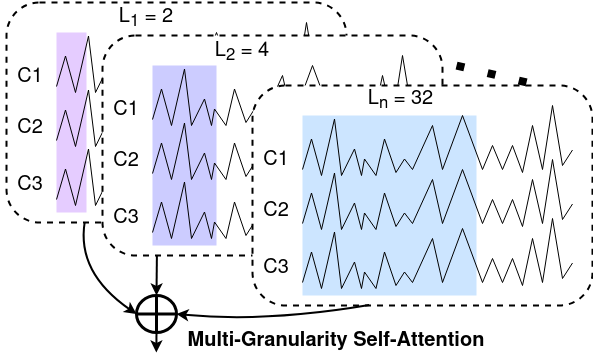}
    \captionof{figure}{multi-Granularity workflow}
    \label{fig:mul_granu}
\end{minipage}

\vspace{0.05cm}

Transformer \cite{vaswani2017attention} and Medformer \cite{wang2024medformermultigranularitypatchingtransformer} explore cross-channel or patch-based representations. PatchTST struggles with cross-channel dependencies, while Medformer addresses this but suffers from information loss and limited feature extraction. Despite advancements, capturing cross-channel dependencies while preserving high-resolution information remains a challenge. To address these, we propose \textbf{Cardioformer}, a multi-granularity hybrid model that integrates cross-channel patching, hierarchical residual learning, and a two-stage self-attention mechanism. It encodes multi-scale token embeddings to capture both local and global ECG features, with self-attention enhancing feature extraction and mitigating tokenization-induced information loss.

We evaluate Cardioformer on MIMIC-IV, PTB-XL, and PTB under subject-independent settings, comparing it to PatchTST, Reformer, Transformer, and Medformer. Cardioformer achieves superior performance with an Area Under the Receiver Operating Characteristic Curve (AUROC) of 96.34\%, F1-score of 81.49\%, and Area Under the Precision-Recall Curve (AUPRC) of 87.65 on MIMIC-IV. These results validate its multi-granularity, hierarchical extraction approach and highlight its potential for automated ECG analysis and cardiovascular disease diagnosis.

\section{Related Work}
ECG signals are key for cardiovascular diagnosis, with research on time series analysis \cite{xiao2023deep, wang2023hierarchical}. Early methods relied on handcrafted features and traditional machine learning, while deep learning used CNNs and RNNs for modeling ECG characteristics \cite{wu2023timesnet, lu2024cats}. Deep learning has significantly improved ECG classification, with CNN-based models like ResNet capturing spatial dependencies \cite{lawhern2018eegnet, song2022eeg}. However, these models struggle with long-range temporal dependencies, crucial for accurate cardiac signal interpretation. Transformers have gained popularity in ECG analysis for modeling long-range dependencies \cite{vaswani2017attention}. Autoformer \cite{wu2021autoformer} and Informer \cite{zhou2021informer} focus on single-channel data, while iTransformer \cite{liu2024itransformer} and PatchTST \cite{nie2023timeseries} use patch-based representations. Reformer \cite{kitaev2019reformer} and standard Transformer \cite{vaswani2017attention} extend these ideas but face challenges in integrating cross-channel dependencies. Recent works explore multi-granularity learning to address single-scale limitations. Medformer \cite{wang2024medformermultigranularitypatchingtransformer} employs cross-channel patching and two-stage self-attention, while MTST \cite{zhang2024multiresolution} and Pathformer \cite{chen2024pathformer} enhance multi-scale pattern recognition but may not fully leverage inter-lead ECG relationships. Table~\ref{tab:medts_characteristics} summarizes these models, comparing multi-timestamp processing, cross-channel learning, and multi-granularity feature extraction.


\begin{table}[ht]
\centering
\caption{\textbf{Key Characteristics of Existing Time Series Models in ECG Analysis.}}
\label{tab:medts_characteristics}
\renewcommand{\arraystretch}{1.2}
\setlength{\tabcolsep}{2pt}
\resizebox{0.7\textwidth}{!}{
\begin{tabular}{lcccc}
\hline
\textbf{Model} & \textbf{Multi-Timestamp} & \textbf{Cross-Channel} & \textbf{Multi-Granularity} & \textbf{Granularity Interaction} \\
\hline
Autoformer \cite{wu2021autoformer} & $\checkmark$ &  &  &  \\
Informer \cite{zhou2021informer} & $\checkmark$ &  &  &  \\
iTransformer \cite{liu2024itransformer} & $\checkmark$ & $\checkmark$ &  &  \\
MTST \cite{zhang2024multiresolution} & $\checkmark$ & $\checkmark$ &  &  \\
PatchTST \cite{nie2023timeseries} & $\checkmark$ &  &  &  \\
Pathformer \cite{chen2024pathformer} & $\checkmark$ & $\checkmark$ &  &  \\
Reformer \cite{kitaev2019reformer} & $\checkmark$ &  &  &  \\
Transformer \cite{vaswani2017attention} & $\checkmark$ &  &  &  \\
Medformer \cite{wang2024medformermultigranularitypatchingtransformer} & $\checkmark$ & $\checkmark$ & $\checkmark$ & $\checkmark$ \\
\textbf{Cardioformer (Ours)} & $\checkmark$ & $\checkmark$ & $\checkmark$ & $\checkmark$ \\
\hline
\end{tabular}
}
\end{table}

In summary, while traditional deep learning models laid the foundation for ECG classification, transformer-based methods and multi-granularity feature learning have advanced multi-channel ECG analysis. Our Cardioformer integrates a novel hierarchical feature extraction framework within a transformer architecture, enhancing the capture of local and global ECG characteristics.

\section{Method}
\label{sec:methodology}

\begin{figure}[ht]
  \centering
  \ifpdf
    \includegraphics[scale=0.085]{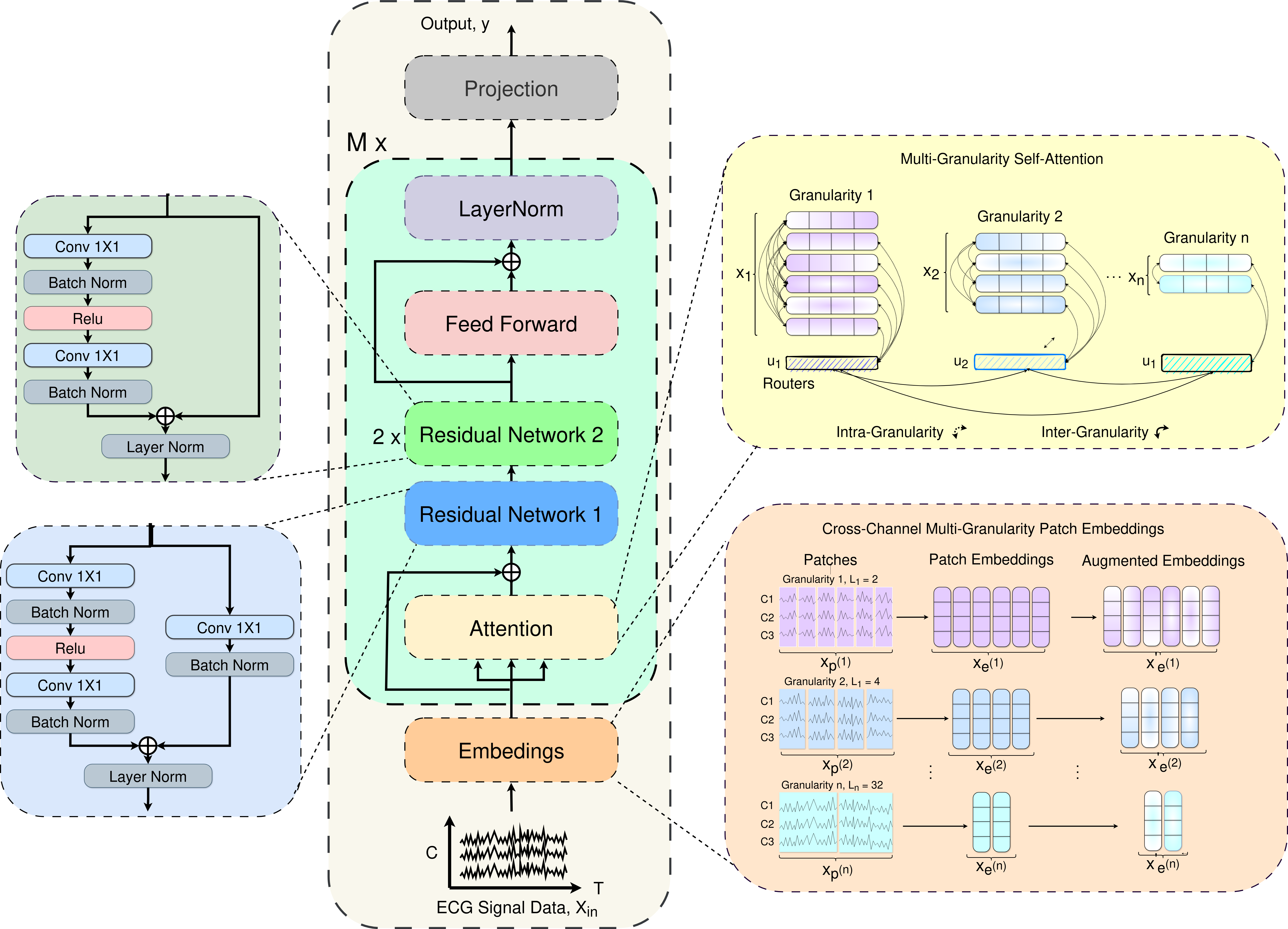}
  \else
    \includegraphics[scale=0.6,natwidth=330,natheight=120]{figs/sample_fig.png}
  \fi
  \caption{\textbf{Architecture of Cardioformer for ECG classification, integrating multi-granularity embeddings, self-attention, and residual networks for enhanced feature extraction and classification.}}
  \label{fig:model_arc}
\end{figure}

We propose Cardioformer for learning spatiotemporal features from multivariate ECG signals, combining cross-channel multi-granularity patch embedding with a two-stage self-attention mechanism, as shown in Figure \ref{fig:model_arc}.

\textbf{(1) Cross-Channel Multi-Granularity Patch Embedding}: ECG channels represent distinct heart regions with inherent correlations. To capture these, we introduce cross-channel multi-granularity patch embedding. Given input $x_{in} \in \mathbb{R}^{T \times C}$, it is divided into non-overlapping patches of lengths $\{L_1, L_2, \dots, L_n\}$ corresponding to different granularities, mapped into latent embeddings via linear projection:
\[
x^{(i)}_e = x^{(i)}_p W^{(i)}, \quad x^{(i)}_e \in \mathbb{R}^{N_i \times D}, \quad W^{(i)} \in \mathbb{R}^{(L_i \cdot C) \times D}.
\]
We apply data augmentation, such as masking and jittering, to improve the embeddings for the following multi-granularity attention stage. Fixed positional embedding $W_{pos} \in \mathbb{R}^{G \times D}$ and learnable granularity embedding $W^{(i)}_{gr}$ preserve positional and granularity distinctions:
\[
x^{(i)} = x^{(i)}_e + W_{pos}[1 : N_i] + W^{(i)}_{gr}, \quad x^{(i)} \in \mathbb{R}^{N_i \times D}.
\]
Router embedding $u^{(i)}$ per granularity:
\[
u^{(i)} = W_{pos}[N_i + 1] + W^{(i)}_{gr}.
\]
The final patch embeddings $\{ x^{(1)}, x^{(2)}, \dots, x^{(n)} \}$ and router embeddings $\{ u^{(1)}, u^{(2)}, \dots, u^{(n)} \}$ are then fed into the multi-granularity self-attention mechanism.

\textbf{(2) Multi-Granularity Self-Attention}: The self-attention mechanism \cite{vaswani2017attention} consists of intra- and inter-granularity stages \cite{wang2024medformermultigranularitypatchingtransformer}. Instead of concatenating all patch embeddings, which increases time complexity, we reduce this by splitting the self-attention into two stages. while \textbf{Intra-Granularity Self-Attention} Concatenate patch and router embeddings per granularity:
\[
z^{(i)} = [ x^{(i)} \parallel u^{(i)} ].
\]
Refine embeddings via attention:
\[
x^{(i)} \leftarrow \text{AttnIntra}(x^{(i)}, z^{(i)}, z^{(i)}), \quad u^{(i)} \leftarrow \text{AttnIntra}(u^{(i)}, z^{(i)}, z^{(i)}).
\]

\textbf{Inter-Granularity Self-Attention} Concatenate routers across granularities:
\[
U = [ u^{(1)} \parallel u^{(2)} \parallel \dots \parallel u^{(n)} ].
\]
Cross-granularity attention updates routers:
\[
u^{(i)} \leftarrow \text{AttnInter}(u^{(i)}, U, U).
\]
Complexity reduces from $O\left((\sum N_i)^2\right)$ to $O(nD^2)$.

\textbf{(3) Residual Network}: Residual blocks (Fig.~\ref{fig:model_arc}) use convolutions, batch normalization (BN), and ReLU ($\sigma$):
\[
\mathbf{y} = \mathcal{F}(\mathbf{x}) + \mathbf{x},\quad 
\mathcal{F}(\mathbf{x}) = \text{BN}(\text{Conv}_{1\times1}(\sigma(\text{BN}(\text{Conv}_{1\times1}(\mathbf{x})))))
\]
First block includes additional skip connection:
\[
\mathbf{y} = \mathcal{F}(\mathbf{x}) + \text{BN}(\text{Conv}_{1\times1}(\mathbf{x}))
\]
Final LayerNorm ensures stability:
\[
\mathbf{y} = \text{LayerNorm}(\mathbf{y})
\]
Stacking $3$ residual blocks enhances features. Our method employs the standard Transformer architecture (Figure \ref{fig:model_arc}). Given an input sample \( \mathbf{x}_{\text{in}} \), it undergoes \( M \) layers of self-attention, producing updated patch embeddings \( \mathbf{x}^{(1)}, \mathbf{x}^{(2)}, \dots, \mathbf{x}^{(n)} \). These are concatenated to form the final representation \( \mathbf{h} \), which is used to predict the label \( y \in \mathbb{R}^{K} \) in a classification task.

\section{Experiments}
\label{sec:expri}

We evaluate Cardioformer against four baseline models across three ECG datasets using a subject-independent split for training, validation, and testing. The datasets include: (1) PTB~\cite{PhysioNet} with binary labels for Myocardial Infarction, (2) PTB-XL~\cite{wagner2022ptbxl} with five-class labels, and (3) MIMIC-IV~\cite{MIMIC-IV-ECG} with four-class labels (see Appendix~\ref{appendix:preprocessing} for preprocessing details). Table~\ref{tab:processed_datasets} summarizes dataset statistics.

\begin{table}[h]
    \centering
    \renewcommand{\arraystretch}{1.2}
    \setlength{\tabcolsep}{3pt} 
    \resizebox{0.8\textwidth}{!}{ 
    \begin{tabular}{>{\centering\arraybackslash}p{1.5cm}
                    >{\centering\arraybackslash}p{1.2cm}
                    >{\centering\arraybackslash}p{1.2cm}
                    >{\centering\arraybackslash}p{1cm}
                    >{\centering\arraybackslash}p{1.4cm}
                    >{\centering\arraybackslash}p{1.9cm}
                    >{\centering\arraybackslash}p{2.9cm}
                    >{\centering\arraybackslash}p{1.5cm}}
        \hline
        \textbf{Datasets} & \textbf{Subject} & \textbf{Sample} & \textbf{Class} & \textbf{Channel} & \textbf{Timestamps} & \textbf{Sampling Rate} &  \textbf{File Size} \\
        \hline
        PTB     & 198    & 64,356  & 2  & 15  & 300  & 250Hz  & 2.32GB \\
        PTB-XL  & 17,596 & 191,400 & 5  & 12  & 250  & 250Hz   & 4.60GB \\
        MIMIC-IV  & 19,931 & 199,310 & 4  & 12  & 250  & 250Hz   & 4.79GB \\
        \hline
    \end{tabular}
    }
    \caption{\textbf{Information of processed datasets, including the number of subjects, samples, classes, channels, sampling rate, sample timestamps, and file size.}}
    \label{tab:processed_datasets}
\end{table}

\textbf{Results}: We compare Cardioformer with four state-of-the-art time series transformer models: PatchTST~\cite{nie2023timeseries}, Reformer~\cite{kitaev2019reformer}, Transformer~\cite{liu2024itransformer}, and Medformer~\cite{wang2024medformermultigranularitypatchingtransformer}. All models are trained on Kaggle P100 GPUs with three random seeds using six encoder layers, three ResNet blocks, 128-dimensional embeddings, and 256-dimensional feed-forward layers. We use the Adam optimizer (learning rate 1e-4), batch sizes of 32 (PTB) and 16 (PTB-XL, MIMIC-IV), and apply early stopping after 3 epochs without improvement in validation F1. The best model checkpoint is used for testing. Data augmentation and additional implementation details are provided in Appendix~\ref{appendix:implementation}.

Table \ref{tab:experiment_results} presents the performance of Cardioformer and the baseline models across three datasets. Cardioformer achieves the highest performance in all metrics on PTB, PTB-XL, and MIMIC-IV. Specifically, on PTB, it achieves 92.84\% accuracy and 95.59\% AUROC, surpassing all baselines. On PTB-XL, Cardioformer also outperforms other models with an accuracy of 73.43\% and an AUROC of 89.99\%. On MIMIC-IV, Cardioformer maintains the best performance with 84.71\% accuracy and 96.34\% AUROC.

\begin{table}[h]
    \centering
    \small
    \renewcommand{\arraystretch}{1.1}
    \setlength{\tabcolsep}{3.5pt} 
    \resizebox{0.85\textwidth}{!}{ 
    \begin{tabular}{>{\centering\arraybackslash}p{1.7cm}
                    >{\raggedright\arraybackslash}p{2cm} 
                    >{\centering\arraybackslash}p{1.4cm} 
                    >{\centering\arraybackslash}p{1.4cm} 
                    >{\centering\arraybackslash}p{1.4cm} 
                    >{\centering\arraybackslash}p{1.4cm} 
                    >{\centering\arraybackslash}p{1.4cm} 
                    >{\centering\arraybackslash}p{1.4cm}} 
        \hline
        \textbf{Datasets} & \textbf{Models} & \textbf{Accuracy} & \textbf{Precision} & \textbf{Recall} & \textbf{F1 Score} & \textbf{AUROC} & \textbf{AUPRC} \\
        \hline
        \multirow{5}{*}{\parbox{1.5cm}{\centering \textbf{ PTB } (2-classes)}} 
        & \textbf{PatchTST} & 91.93$\pm$\text{\tiny 0.55} & 84.98$\pm$\text{\tiny 1.38} & 81.63$\pm$\text{\tiny 1.81} & 83.17$\pm$\text{\tiny 1.50} & 93.85$\pm$\text{\tiny 1.36} & 87.37$\pm$\text{\tiny 2.09} \\
        & \textbf{Reformer} & 93.87$\pm$\text{\tiny 0.27} & \textbf{88.15$\pm$\text{\tiny 0.75}} & 86.35$\pm$\text{\tiny 2.37} & 85.68$\pm$\text{\tiny 0.97} & 93.03$\pm$\text{\tiny 0.62} & 87.75$\pm$\text{\tiny 0.95} \\
        & \textbf{Transformer} & 92.79$\pm$\text{\tiny 0.69} & 85.02$\pm$\text{\tiny 2.00} & 83.33$\pm$\text{\tiny 2.78} & 84.92$\pm$\text{\tiny 1.67} & 94.34$\pm$\text{\tiny 0.98} & 88.69$\pm$\text{\tiny 2.07} \\
        & \textbf{Medformer} & 92.13$\pm$\text{\tiny 0.23} & 86.81$\pm$\text{\tiny 3.42} & 81.50$\pm$\text{\tiny 4.76} & 83.15$\pm$\text{\tiny 2.12} & 94.49$\pm$\text{\tiny 0.36} & 88.96$\pm$\text{\tiny 0.98} \\
        & \textbf{Cardioformer} (Ours) & \textbf{92.84$\pm$\text{\tiny 0.92}} & 85.23$\pm$\text{\tiny 1.59} & \textbf{87.09$\pm$\text{\tiny 2.65}} & \textbf{86.11$\pm$\text{\tiny 1.99}} & \textbf{95.59$\pm$\text{\tiny 1.66}} & \textbf{89.50$\pm$\text{\tiny 4.13}} \\
        \hline
        \multirow{5}{*}{\parbox{1.5cm}{\centering \textbf{ PTB-XL } (5-classes)}} 
        & \textbf{PatchTST} & 71.95$\pm$\text{\tiny 0.57} & \textbf{68.09$\pm$\text{\tiny 1.41}} & 56.03$\pm$\text{\tiny 1.72} & 56.52$\pm$\text{\tiny 1.44} & 88.79$\pm$\text{\tiny 0.25} & 65.94$\pm$\text{\tiny 0.6} \\     
        & \textbf{Reformer} & 72.40$\pm$\text{\tiny 0.07} & 65.32$\pm$\text{\tiny 0.57} & 56.50$\pm$\text{\tiny 0.37} & 57.48$\pm$\text{\tiny 0.08} & 88.44$\pm$\text{\tiny 0.12} & 65.42$\pm$\text{\tiny 0.15} \\     
        & \textbf{Transformer} & 71.16$\pm$\text{\tiny 0.33} & 62.94$\pm$\text{\tiny 0.50} & 57.53$\pm$\text{\tiny 0.09} & 59.25$\pm$\text{\tiny 0.08} & 87.74$\pm$\text{\tiny 0.13} & 63.71$\pm$\text{\tiny 0.13} \\
        & \textbf{Medformer} & 72.51$\pm$\text{\tiny 0.19} & 64.46$\pm$\text{\tiny 0.44} & 60.04$\pm$\text{\tiny 0.20} & 61.72$\pm$\text{\tiny 0.35} & 88.93$\pm$\text{\tiny 0.17} & 66.13$\pm$\text{\tiny 0.40} \\

        & \textbf{Cardioformer} (Ours) & \textbf{73.43$\pm$\text{\tiny 0.46}} & 65.28$\pm$\text{\tiny 0.73} & \textbf{60.27$\pm$\text{\tiny 0.10}} & \textbf{62.07$\pm$\text{\tiny 0.29}} & \textbf{89.99$\pm$\text{\tiny 0.12}} & \textbf{67.07$\pm$\text{\tiny 0.40}} \\
        \hline
        \multirow{5}{*}{\parbox{1.7cm}{\centering \textbf{ MIMIC-IV } (4-classes)}} 
        & \textbf{PatchTST} & 85.00$\pm$\text{\tiny 0.10} & \textbf{82.17$\pm$\text{\tiny 0.24}} & \textbf{82.88$\pm$\text{\tiny 0.41}} & 79.51$\pm$\text{\tiny 0.20} & 95.86$\pm$\text{\tiny 0.05} & 86.56$\pm$\text{\tiny 0.23} \\
        & \textbf{Reformer} & 80.90$\pm$\text{\tiny 0.51} & 80.03$\pm$\text{\tiny 0.6} & 74.64$\pm$\text{\tiny 0.8} & 76.80$\pm$\text{\tiny 0.73} & 94.66$\pm$\text{\tiny 0.15} & 84.34$\pm$\text{\tiny 0.54} \\
        & \textbf{Transformer} & 82.92$\pm$\text{\tiny 0.50} & 79.69$\pm$\text{\tiny 0.21} & 79.30$\pm$\text{\tiny 1.37} & 79.34$\pm$\text{\tiny 0.96} & 95.05$\pm$\text{\tiny 0.12} & 85.48$\pm$\text{\tiny 0.47} \\

        & \textbf{Medformer} & 83.13$\pm$\text{\tiny 0.18} & 80.38$\pm$\text{\tiny 0.13} & 80.15$\pm$\text{\tiny 0.53} & 80.01$\pm$\text{\tiny 0.26} & 95.76$\pm$\text{\tiny 0.12} & 85.91$\pm$\text{\tiny 0.40} \\
        
        & \textbf{Cardioformer} (Ours) & \textbf{84.71$\pm$\text{\tiny 0.24}} & 81.96$\pm$\text{\tiny 0.28} & 81.25$\pm$\text{\tiny 0.51} & \textbf{81.49$\pm$\text{\tiny 0.42}} & \textbf{96.34$\pm$\text{\tiny 0.11}} & \textbf{87.65$\pm$\text{\tiny 0.42}} \\
        \Xhline{2\arrayrulewidth} 
    \end{tabular}
    }
    \caption{Performance of \textbf{Cardioformer} and baselines on \textbf{PTB}, \textbf{PTB-XL}, and \textbf{MIMIC-IV} datasets.}
    \label{tab:experiment_results}
\end{table}

Table \ref{tab:cross_result} shows Cardioformer’s cross-dataset generalization. Trained solely on MIMIC-IV, it performs well on PTB (51.13\% accuracy, 49.18\% AUROC) and PTB-XL (62.28\% accuracy, 68.41\% AUROC), highlighting Cardioformer’s robustness when transferred to unseen datasets, underscoring its adaptability in real-world applications.

\begin{table}[h]
    \centering
    \small
    \renewcommand{\arraystretch}{1.25}
    \setlength{\tabcolsep}{3.5pt}
    \resizebox{0.8\textwidth}{!}{ 
    \begin{tabular}{>{\centering\arraybackslash}p{2cm}
                    >{\centering\arraybackslash}p{1.7cm} 
                    >{\centering\arraybackslash}p{1.4cm} 
                    >{\centering\arraybackslash}p{1.4cm} 
                    >{\centering\arraybackslash}p{1.4cm} 
                    >{\centering\arraybackslash}p{1.4cm} 
                    >{\centering\arraybackslash}p{1.4cm} 
                    >{\centering\arraybackslash}p{1.4cm}} 
        \hline
        \textbf{Model} & \textbf{Dataset} & \textbf{Accuracy} & \textbf{Precision} & \textbf{Recall} & \textbf{F1 Score} & \textbf{AUROC} & \textbf{AUPRC} \\
        \hline
        \multirow{3}{*}{\textbf{PatchTST}} 
        & \textbf{PTB} & 60.55$\pm$\text{\tiny 31.14} & 43.11$\pm$\text{\tiny 0.01} & 50.00$\pm$\text{\tiny 0.23} & 45.26$\pm$\text{\tiny 11.41} & 47.26$\pm$\text{\tiny 5.01} & 49.63$\pm$\text{\tiny 3.35} \\
        & \textbf{PTB-XL} & 57.60$\pm$\text{\tiny 0.16} & 59.35$\pm$\text{\tiny 0.38} & 58.83$\pm$\text{\tiny 0.30} & 57.32$\pm$\text{\tiny 0.24} & 63.84$\pm$\text{\tiny 0.07} & 62.14$\pm$\text{\tiny 0.07} \\
        & \textbf{MIMIC-IV} & 82.64$\pm$\text{\tiny 0.03} & 82.69$\pm$\text{\tiny 0.35} & 82.94$\pm$\text{\tiny 0.24} & 83.63$\pm$\text{\tiny 0.29} & 88.29$\pm$\text{\tiny 0.02} & 88.91$\pm$\text{\tiny 0.39} \\
        \hline
        \multirow{3}{*}{\textbf{Reformer}} 
        & \textbf{PTB} & 60.04$\pm$\text{\tiny 7.53} & 51.44$\pm$\text{\tiny 5.65} & 53.31$\pm$\text{\tiny 6.39} & 50.62$\pm$\text{\tiny 5.25} & 52.05$\pm$\text{\tiny 11.29} & 51.91$\pm$\text{\tiny 5.13} \\
        & \textbf{PTB-XL} & 55.40$\pm$\text{\tiny 0.33} & 59.57$\pm$\text{\tiny 0.09} & 57.59$\pm$\text{\tiny 0.08} & 53.80$\pm$\text{\tiny 0.08} & 64.43$\pm$\text{\tiny 0.13} & 62.47$\pm$\text{\tiny 0.21} \\
        & \textbf{MIMIC-IV} & 83.66$\pm$\text{\tiny 0.18} & 83.92$\pm$\text{\tiny 0.12} & 83.26$\pm$\text{\tiny 0.25} & 82.47$\pm$\text{\tiny 0.06} & 91.21$\pm$\text{\tiny 0.12} & 90.86$\pm$\text{\tiny 0.15} \\
        \hline
        \multirow{3}{*}{\textbf{Transformer}} 
        & \textbf{PTB} & 69.64$\pm$\text{\tiny 25.29} & 45.30$\pm$\text{\tiny 4.31} & 47.92$\pm$\text{\tiny 8.61} & 45.05$\pm$\text{\tiny 12.72} & 44.70$\pm$\text{\tiny 15.11} & 49.08$\pm$\text{\tiny 5.36} \\
        & \textbf{PTB-XL} & 56.05$\pm$\text{\tiny 0.19} & 58.94$\pm$\text{\tiny 0.31} & 57.80$\pm$\text{\tiny 0.36} & 55.21$\pm$\text{\tiny 0.20} & 62.44$\pm$\text{\tiny 0.29} & 60.79$\pm$\text{\tiny 0.14} \\
        & \textbf{MIMIC-IV} & 85.31$\pm$\text{\tiny 0.11} & 85.23$\pm$\text{\tiny 0.08} & 85.43$\pm$\text{\tiny 0.13} & 85.27$\pm$\text{\tiny 0.22} & 92.17$\pm$\text{\tiny 0.26} & 91.70$\pm$\text{\tiny 0.17} \\
        \hline
        \multirow{3}{*}{\textbf{Medformer}} 
        & \textbf{PTB} & 52.11$\pm$\text{\tiny 18.58} & 49.40$\pm$\text{\tiny 3.24} & 50.88$\pm$\text{\tiny 2.79} & 39.89$\pm$\text{\tiny 4.66} & 51.69$\pm$\text{\tiny 4.62} & 50.95$\pm$\text{\tiny 1.64} \\
        & \textbf{PTB-XL} & 54.81$\pm$\text{\tiny 0.28} & 58.11$\pm$\text{\tiny 0.32} & 56.77$\pm$\text{\tiny 0.30} & 53.59$\pm$\text{\tiny 0.15} & 62.99$\pm$\text{\tiny 0.19} & 60.87$\pm$\text{\tiny 0.27} \\
        & \textbf{MIMIC-IV} & 86.93$\pm$\text{\tiny 0.36} & 87.03$\pm$\text{\tiny 0.20} & 86.64$\pm$\text{\tiny 0.38} & 86.78$\pm$\text{\tiny 0.09} & 92.18$\pm$\text{\tiny 0.25} & 91.69$\pm$\text{\tiny 0.16} \\
        \hline
        \multirow{3}{*}{\textbf{Cardioformer}} 
        & \textbf{PTB} & 51.13$\pm$\text{\tiny 4.65} & 46.72$\pm$\text{\tiny 3.53} & 48.01$\pm$\text{\tiny 0.87} & 42.78$\pm$\text{\tiny 7.39} & 49.18$\pm$\text{\tiny 4.62} & 49.73$\pm$\text{\tiny 3.51} \\
        & \textbf{PTB-XL} & 62.28$\pm$\text{\tiny 1.77} & 64.75$\pm$\text{\tiny 1.10} & 60.23$\pm$\text{\tiny 3.18} & 57.70$\pm$\text{\tiny 4.98} & 68.41$\pm$\text{\tiny 2.69} & 66.53$\pm$\text{\tiny 2.76} \\
        & \textbf{MIMIC-IV} & 88.28$\pm$\text{\tiny 0.39} & 88.27$\pm$\text{\tiny 0.15} & 88.12$\pm$\text{\tiny 0.24} & 88.19$\pm$\text{\tiny 0.13} & 94.91$\pm$\text{\tiny 0.05} & 94.33$\pm$\text{\tiny 0.01} \\
        \Xhline{2\arrayrulewidth}
    \end{tabular}
    }
    \caption{Cross-dataset generalization performance of \textbf{Cardioformer} and baseline models on the \textbf{PTB}, \textbf{PTB-XL}, and \textbf{MIMIC-IV} datasets, where all models are trained exclusively on the \textbf{MIMIC-IV} dataset.}
    \label{tab:cross_result}
\end{table}

\section{Conclusion and Limitations}

In this paper, we introduced \textbf{Cardioformer}, a novel multi-granularity patching transformer with a ResNet block for ECG classification. Cardioformer effectively captures multi-timestamp and cross-channel features through multi-granularity token embeddings and a two-stage self-attention mechanism, extracting both local morphological details and long-range temporal dependencies. Experimental evaluations on three benchmark datasets show that Cardioformer outperforms state-of-the-art baselines, demonstrating its potential for real-world applications and cross-dataset generalization. However, several limitations remain. First, the model's flexibility with variable patch lengths introduces hyperparameter sensitivity. Future work could explore adaptive strategies for selecting optimal patch lengths. Second, although the two-stage self-attention integrates multi-granularity features, it doesn’t fully account for the varying importance of ECG channels. Selective channel attention could improve accuracy, particularly for subtle abnormality detection. Finally, integrating selective state space models (SSMs) could enhance temporal modeling, improving heart disease detection. Addressing these limitations will refine Cardioformer and advance automated ECG analysis for cardiovascular disease diagnosis.


\printbibliography[heading=subbibintoc]

\appendix

\section{Data Augmentation Techniques}
\label{appendix:augment_details}

In the embedding stage, we incorporate a set of data augmentation strategies to enhance the robustness and generalization capability of the model. During each forward pass in training, a patch embedding is augmented using one randomly selected technique from the augmentation pool, with all options having equal selection probability. The augmentation methods used in this work are as follows:

\textbf{Channel Shuffling:} The order of ECG channels is randomly shuffled. This augmentation is applied with a default probability of 0.5, controlled by the parameter \texttt{prob}.

\textbf{Temporal Masking:} A random subset of timestamps across all channels is masked. The proportion of masked timestamps is determined by the parameter \texttt{ratio}, with a default value of 0.1.

\textbf{Frequency Masking:} The ECG signal is first transformed into the frequency domain, after which a portion of frequency bands is masked, followed by an inverse transformation. The masked proportion is governed by \texttt{ratio = 0.1}.

\textbf{Jittering:} Random noise is added to the raw ECG signal. The noise level is scaled by the parameter \texttt{scale}, which is set to 0.1 by default.

\textbf{Dropout:} Similar to dropout in neural networks, this technique randomly omits certain values from the input signal. The dropout rate is specified by the parameter \texttt{ratio}, with a default setting of 0.1.

These augmentation methods are modular and can be extended to include additional strategies as needed.

\section{Data Preprocessing}
\label{appendix:preprocessing}
\subsection{PTB Preprocessing}
The PTB dataset \cite{PhysioNet} comprises multi-channel ECG recordings from 290 subjects, each annotated with one of eight diagnostic categories—seven representing different cardiac pathologies and one denoting healthy control. The recordings are sampled at 1000\,Hz across 15 leads. In this study, we consider a subset of 198 subjects, including only those diagnosed with myocardial infarction or labeled as healthy.

To standardize the input, signals are downsampled to 250\,Hz and normalized using a standard scaler. Individual heartbeats are extracted by identifying R-peaks across all channels, followed by the removal of outliers. Each heartbeat is sampled around its corresponding R-peak, and zero-padding is applied to ensure consistent length, using the maximum duration observed across all beats as reference. This procedure yields a total of 64,356 heartbeat samples. We adopt a subject-independent split, assigning 60\% of subjects to the training set, 20\% to validation, and 20\% to testing.

\subsection{PTB-XL Preprocessing}
The PTB-XL dataset \cite{wagner2022ptbxl} is a large-scale ECG database containing 12-lead recordings from 18,869 subjects, labeled across five diagnostic categories—four corresponding to cardiac abnormalities and one to healthy controls. To ensure label consistency, we exclude subjects exhibiting divergent diagnostic labels across multiple trials, retaining a cohort of 17,596 subjects.

We utilize the 500\,Hz version of the recordings, which are subsequently downsampled to 250\,Hz and standardized using a normal scaler. Each 10-second trial is segmented into non-overlapping 1-second intervals comprising 250 time points. Segments shorter than one second are discarded, resulting in 191,400 samples. A subject-independent data partitioning strategy is employed, allocating 60\% of subjects to the training set, 20\% to validation, and 20\% to testing.

\subsection{MIMIC-IV Preprocessing}
The MIMIC-IV-ECG dataset \cite{MIMIC-IV-ECG} comprises approximately 800,000 diagnostic 12-lead ECG recordings from nearly 160,000 unique patients, collected between 2008 and 2019 at the Beth Israel Deaconess Medical Center. Each recording is 10 seconds in duration and sampled at 500\,Hz. These ECGs are linked to the MIMIC-IV Clinical Database, facilitating access to associated clinical information.

For this study, we selected a subset of ECGs corresponding to patients with consistent diagnostic labels across their records. The ECG signals were downsampled to 250\,Hz and normalized using a standard scaler. R-peaks were detected across all leads to segment the continuous recordings into individual heartbeats. Outliers and noisy segments were excluded during preprocessing. Each heartbeat was centered around its R-peak, and zero-padding was applied to ensure uniform length, based on the maximum duration observed across all samples.

From this preprocessing pipeline, we sampled and utilized a total of 19,931 heartbeat segments. A subject-independent split was employed, assigning 60\% of the subjects to the training set, 20\% to the validation set, and the remaining 20\% to the test set.

\section{Implementation Details}
\label{appendix:implementation}
We implement our method and all the baselines based on the Time-Series-Library project\footnote{\url{https://github.com/thuml/Time-Series-Library}} from Tsinghua University \cite{wu2023timesnet}, which integrates all methods under the same framework and training techniques to ensure a relatively fair comparison. The 4 baseline time series transformer methods are PatchTST \cite{nie2023timeseries}, Reformer \cite{kitaev2019reformer}, Transformer \cite{liu2024itransformer} and Medformer \cite{wang2024medformermultigranularitypatchingtransformer}. 

For all methods, we use 6 encoder layers, with self-attention dimension $D=128$ and feed-forward hidden dimension 256. The Adam optimizer is used with a learning rate of 1e-4. Batch sizes are set to \{32, 16, 16\} for PTB, PTB-XL, and MIMIC-IV, respectively. Models are trained for 10 epochs, with early stopping after 3 epochs without F1 improvement on the validation set. The best validation F1 model is saved and evaluated on the test set. We report six metrics: accuracy, macro-averaged precision, recall, F1 score, AUROC, and AUPRC. Each experiment is run with three random seeds (41–43) using fixed data splits to compute average results and standard deviations.

\textbf{Cardioformer (Our Method)} We use a list of patch lengths in patch embedding to generate patches with different granularities. Instead of flattening the patches and mapping them to dimension D during patch embedding, we used Residual network to directly map patches into a 1-D representation with dimension D. These patch lengths can vary, including different numbers of patch lengths such as \{2, 4, 8, 16\}, repetitive numbers such as \{8, 8, 8, 8\}, or a mix of different and repetitive lengths such as \{8, 8, 8, 16, 16, 16\}. It is also possible to use only one patch length, such as {8}, which indicates a single granularity. The patch lists used for all the datasets PTB, PTB-XL and MIMIC-IV are \{2, 4, 8, 8, 16, 16, 16, 16, 32, 32, 32, 32, 32, 32, 32, 32\}. The data augmentations are randomly chosen from a list of four possible options: none, jitter, scale, and mask. The number following each augmentation method indicates the degree of augmentation. Detailed descriptions of these methods can be found in Appendix ~\ref{appendix:augment_details}. The augmentation methods used for the datasets PTB, PTB-XL and MIMIC-IV are \{drop0.5\}, \{jitter0.2, scale0.2, drop0.5\} and \{jitter0.2, scale0.2, drop0.5\}, respectively.

\end{document}